\documentclass[letterpaper, 10 pt, conference]{ieeeconf}  % Comment this line out if you need a4paper

\IEEEoverridecommandlockouts                              % This command is only needed if 
\usepackage{amsmath,amsfonts}
\usepackage{amssymb}

\usepackage{algorithm}
\usepackage[noend]{algorithmic}
\usepackage{color}

\usepackage{array}
\usepackage{textcomp}
\usepackage{stfloats}

\usepackage{verbatim}
\usepackage{graphicx}

\usepackage{cite}
\usepackage[utf8]{inputenc} % allow utf-8 input
\usepackage[T1]{fontenc}    % use 8-bit T1 fonts
\usepackage{hyperref}       % hyperlinks
\usepackage{url}            % simple URL typesetting
\usepackage{booktabs}       % professional-quality tables
\usepackage{amsfonts}       % blackboard math symbols
\usepackage{nicefrac}       % compact symbols for 1/2, etc.
\usepackage{microtype}      % microtypography
\usepackage{lipsum}
\usepackage{fancyhdr}       % header
\usepackage{graphicx}       % graphics
\usepackage{wrapfig}
\usepackage{subfigure}
\usepackage{multirow}

\usepackage{amsthm}

\usepackage{threeparttable}
\usepackage{makecell}

\title{\LARGE \bf
Fair Play in the Fast Lane: \\ Integrating Sportsmanship into Autonomous Racing Systems
}

\author{Zhenmin Huang$^{1,2*}$, Ce Hao$^{3*}$, Wei Zhan$^2$, Jun Ma$^1$, and Masayoshi Tomizuka$^2$, \textit{Life Fellow, IEEE}
\thanks{$^*$Equally contributed.}
\thanks{$^{1}$Division of Emerging Interdisciplinary Areas, The Hong Kong University of Science and Technology, Hong Kong SAR, China.
        {\tt\small zhuangdf@connect.ust.hk, jun.ma@ust.hk}}%
\thanks{$^{2}$Department of Mechanical Engineering, University of California, Berkeley, CA, USA. {\tt\small \{wzhan, tomizuka\}@berkeley.edu}}
\thanks{$^{3}$School of Computing, National University of Singapore, Singapore.  {\tt\small cehao@u.nus.edu}}%
}

\begin{document}

\theoremstyle{definition}
\newtheorem{definition}{Definition}

\maketitle
\thispagestyle{empty}
\pagestyle{empty}

%%%%%%%%%%%%%%%%%%%%%%%%%%%%%%%%%%%%%%%%%%%%%%%%%%%%%%%%%%%%%%%%%%%%%%%%%%%%%%%%

\begin{abstract}
Autonomous racing has gained significant attention as a platform for high-speed decision-making and motion control. While existing methods primarily focus on trajectory planning and overtaking strategies, the role of sportsmanship in ensuring fair competition remains largely unexplored. In human racing, rules such as the one-motion rule and the enough-space rule prevent dangerous and unsportsmanlike behavior. However, autonomous racing systems often lack mechanisms to enforce these principles, potentially leading to unsafe maneuvers. 
This paper introduces a bi-level game-theoretic framework to integrate sportsmanship (SPS) into versus racing. At the high level, we model racing intentions using a Stackelberg game, where Monte Carlo Tree Search (MCTS) is employed to derive optimal strategies. At the low level, vehicle interactions are formulated as a Generalized Nash Equilibrium Problem (GNEP), ensuring that all agents follow sportsmanship constraints while optimizing their trajectories. 
Simulation results demonstrate the effectiveness of the proposed approach in enforcing sportsmanship rules while maintaining competitive performance. We analyze different scenarios where attackers and defenders adhere to or disregard sportsmanship rules and show how knowledge of these constraints influences strategic decision-making. This work highlights the importance of balancing competition and fairness in autonomous racing and provides a foundation for developing ethical and safe AI-driven racing systems.
\end{abstract}

\section{Introduction} \label{Sec: intro}

Autonomous racing, derived from urban autonomous driving, pushes AI agents to operate at the limits of vehicle dynamics. It serves as a testbed for high-speed decision-making, with competitions evaluating AI control strategies~\cite{betz2022autonomous}.
There are two primary formats: time-trial racing and versus racing. In time-trial racing, vehicles drive separately, and then attend to optimize trajectories to minimize lap time~\cite{hao2022outracing}. Traditional approaches use model-based control, while state-of-the-art methods leverage reinforcement learning (RL) to achieve superior performance~\cite{fuchs2021super}.
% Versus racing introduces competitive dynamics, where a defender defends its position while an attacker attempts to overtake. This creates a game-theoretic challenge requiring strategic decision-making~\cite{song2021autonomous}. \red{(Old) Previous research like GT Sophy~\cite{wurman2022outracing} has shown superhuman performance by integrating RL and opponent-aware planning, outperforming human drivers in head-to-head races.}
Previous research, such as GT Sophy~\cite{wurman2022outracing}, has shown superhuman performance by leveraging RL to develop opponent-aware racing behaviors, outperforming human drivers in head-to-head races.

\begin{figure}[t]
    \centering
    \includegraphics[width=0.97\columnwidth]{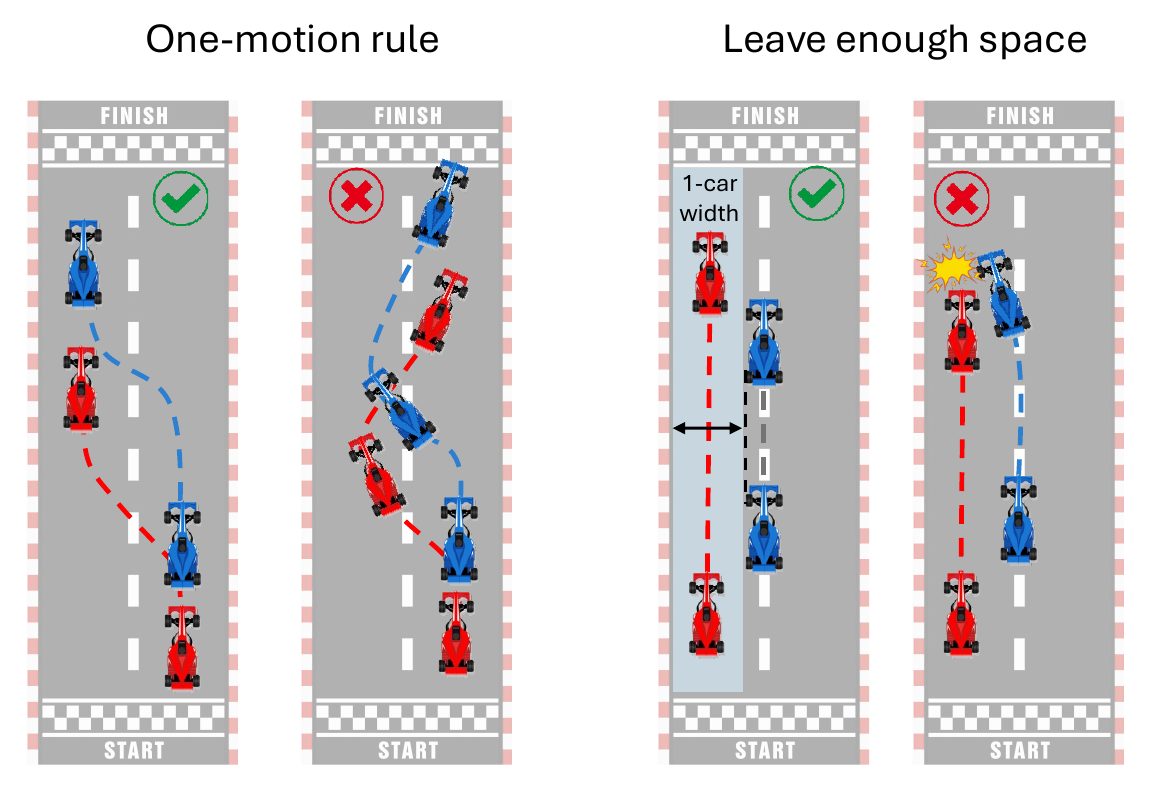}
    \caption{Sportsmanship in autonomous racing. \textbf{Left}: Under the one-motion rule, the defender is allowed to make only a single lateral move to block the attacker. Continuous swerving to defend its position is prohibited. \textbf{Right}: Under the enough-space rule, when the attacker attempts to overtake along the edge of the track boundary, the defender must leave sufficient space. Deliberately cutting in to force the attacker to brake or risk a collision is not permitted.}
    \label{Fig: teaser}
\end{figure}

Beyond securing victory, sportsmanship (SPS) is essential in competitive racing, ensuring fairness and safety. In human-driven races, SPS prevents reckless maneuvers that endanger opponents, such as deliberate collisions or forcing risky actions~\cite{smith2016race}.
As in Fig.~\ref{Fig: teaser}, two key SPS rules are widely enforced. The One-motion rule prohibits a defender from making frequent lateral moves to block an attacker, preventing abrupt turns and potential crashes. The Enough-space rule ensures that when an attacker is near the track boundary, the defender cannot aggressively cut into its path, forcing hard braking. Violating SPS is considered unsportsmanlike and is subject to penalties from race officials.
While SPS is a fundamental principle in human racing, it has been largely overlooked in autonomous racing. Current AI-driven systems focus solely on optimizing speed and overtaking strategies, often disregarding fairness and safety. This absence of SPS enforcement can lead to aggressive, unreasonable, or even dangerous driving behaviors, posing significant risks when deploying autonomous racing systems in real-world applications~\cite{gutenberger2024modeling}. In this sense, ensuring AI respects SPS rules is crucial for safe and ethical autonomous competition.

In this paper, we investigate the role of SPS in autonomous racing and aim to address the following key questions:
1) How can unsportsmanlike behaviors be identified and quantified?
2) Under what conditions is SPS enforcement necessary in racing?
3) How can SPS rules be systematically integrated into autonomous racing algorithms?
4) What impact does SPS have on strategic decision-making in versus racing?
A major challenge in enforcing SPS is that it is inherently non-differentiable and requires long-horizon trajectory analysis to assess compliance. Unlike conventional optimization-based racing strategies that prioritize speed and overtaking efficiency, incorporating SPS involves reasoning about fairness constraints, competitive dynamics, and ethical decision-making. Furthermore, the coupled nature of racing interactions and the nonlinear complexity of vehicle trajectory planning pose additional difficulties in ensuring that autonomous agents adhere to SPS guidelines while maintaining competitive performance.

% To answer these questions is not a trivial task, as SPS is inherently non-differentiable, and the determination of SPS often requires inspecting a long horizon of trajectories. Together with the coupled nature of racing behaviors and the nonlinearity involved in trajectory planning of racing vehicles, these problems present significant challenges.
In this paper, we propose an intention-trajectory planning framework, which effectively incorporates the SPS of racing vehicle competition in versus autonomous racing games through bi-level game formulation. At the high-level game, the intentions of racing vehicles are formulated as a discrete Stackelberg game~\cite{fisac2019hierarchical}, and Monte Carlo Tree Search (MCTS) is applied to obtain near-optimal intention-level strategies~\cite{kocsis2006bandit}. At the low-level game, the interaction between racing vehicles is formulated as a Generalized Nash Equilibrium Problem (GNEP)~\cite{facchinei2010generalized}, which is solved to the corresponding Generalized Nash Equilibrium (GNE) such that the trajectories of all players are obtained with pertinent constraints satisfied. Through bi-level game formulation, we successfully address the aforementioned challenges. Simulation results on different scenarios and different SPS rules verify the proposed method's effectiveness, which also highlight the importance of considering SPS in racing games.

To evaluate the effectiveness of integrating SPS into autonomous racing, we conduct simulations across different racing scenarios, comparing agents that adhere to or disregard SPS principles. We assess performance using key metrics such as overtaking success rates, violation frequencies, and time-to-goal efficiency. The experiments focus on two primary scenarios: straightaways and corners, where attackers and defenders exhibit different levels of SPS awareness. We analyze how these rules influence strategic decision-making, particularly in scenarios where violations lead to penalties or altered race outcomes. By systematically comparing these settings, we establish benchmarks that quantify the impact of fair play on autonomous racing dynamics, demonstrating how SPS enforcement balances competition, fairness, and safety.

In summary, the contributions are threefold:
\begin{enumerate}
    \item We highlight the importance of SPS in autonomous racing and identify the challenges posed by the lack of explicit fairness enforcement in existing AI-driven racing strategies. 
    \item We propose a bi-level game-theoretic framework that integrates SPS principles into versus racing, where high-level decision-making is modeled as a Stackelberg game and low-level trajectory optimization is formulated as GNEP. 
    \item We conduct extensive simulations across various racing scenarios to validate our approach, demonstrating how SPS enforcement influences competitive dynamics and improves fairness while maintaining performance. 
\end{enumerate}

\section{Related Works} \label{Sec: related}

Autonomous racing has emerged as a critical testbed for high-speed decision-making, pushing the boundaries of perception, planning, and control~\cite{betz2022autonomous, fuchs2021super}. Existing research primarily focuses on optimizing vehicle dynamics and trajectory planning for performance-driven objectives~\cite{zhang2024survey, ni2017dynamics}. Traditional methods in autonomous racing leverage model-based approaches such as Model Predictive Control (MPC) to compute the fastest trajectory while maintaining stability~\cite{kabzan2019learning}. Recent advancements integrate RL techniques, demonstrating the capability of learning-based policies to outperform classical methods in specific racing scenarios \cite{williams2019model}. 
% \red{(Old) Studies on GT Sophy and similar AI-driven racers have shown superhuman racing strategies by combining deep RL with opponent-aware planning \cite{wurman2022outracing}. However, these systems primarily optimize for speed and efficiency, often neglecting higher-level behavioral principles such as fairness and SPS.}
Studies on GT Sophy and similar AI-driven racers have demonstrated superhuman racing strategies by utilizing deep RL to develop opponent-aware behaviors, including race car control, tactical decision-making, and racing etiquette \cite{wurman2022outracing}. While GT Sophy has successfully incorporated sportsmanship principles, the broader challenge of systematically enforcing fairness and SPS across diverse autonomous racing scenarios remains an open research question.

While game-theoretic frameworks have been explored to model competitive interactions in autonomous racing, they largely focus on strategic overtaking and defensive maneuvers \cite{liniger2015optimization, song2021autonomous}. Many existing approaches incorporate Nash equilibria and best-response strategies to model competitive interactions, ensuring robust driving policies in adversarial settings~\cite{liniger2019noncooperative, liniger2019noncooperative}. However, these methods do not explicitly account for fairness constraints or SPS enforcement. Unlike human racers, who operate under implicit ethical guidelines and formal race regulations, AI-driven racing agents lack explicit mechanisms to ensure adherence to SPS rules, such as preventing excessive blocking or forcing opponents into unsafe conditions~\cite{smith2016race, simons2003race}. The omission of such principles in autonomous racing models may lead to overly aggressive driving behaviors, which, while optimal for winning, may be unrealistic and unsafe for real-world applications.

Efforts to enforce rule-based constraints in autonomous driving have been explored in broader contexts, such as collision avoidance~\cite{keanly2024trajectory}, traffic law compliance~\cite{maurer2016autonomous}, and safety-critical decision-making \cite{schwarting2019social}. However, these constraints are typically designed for urban driving, where fairness and cooperation are essential for social acceptance. In contrast, the high-speed, competitive nature of autonomous racing presents unique challenges where SPS needs to be balanced with aggressive but fair competitive strategies. To address this gap, our work introduces a bi-level intention-trajectory planning framework that explicitly integrates SPS constraints into competitive autonomous racing. 

% By formulating SPS-aware decision-making as a game-theoretic optimization problem, we aim to bridge the gap between performance-driven AI strategies and human-like ethical considerations in high-speed racing environments.
\section{Preliminary} \label{Sec: prelim}

\subsection{Vehicle Kinematics}

\begin{figure}[h]
    \centering
    \includegraphics[width=0.5\columnwidth]{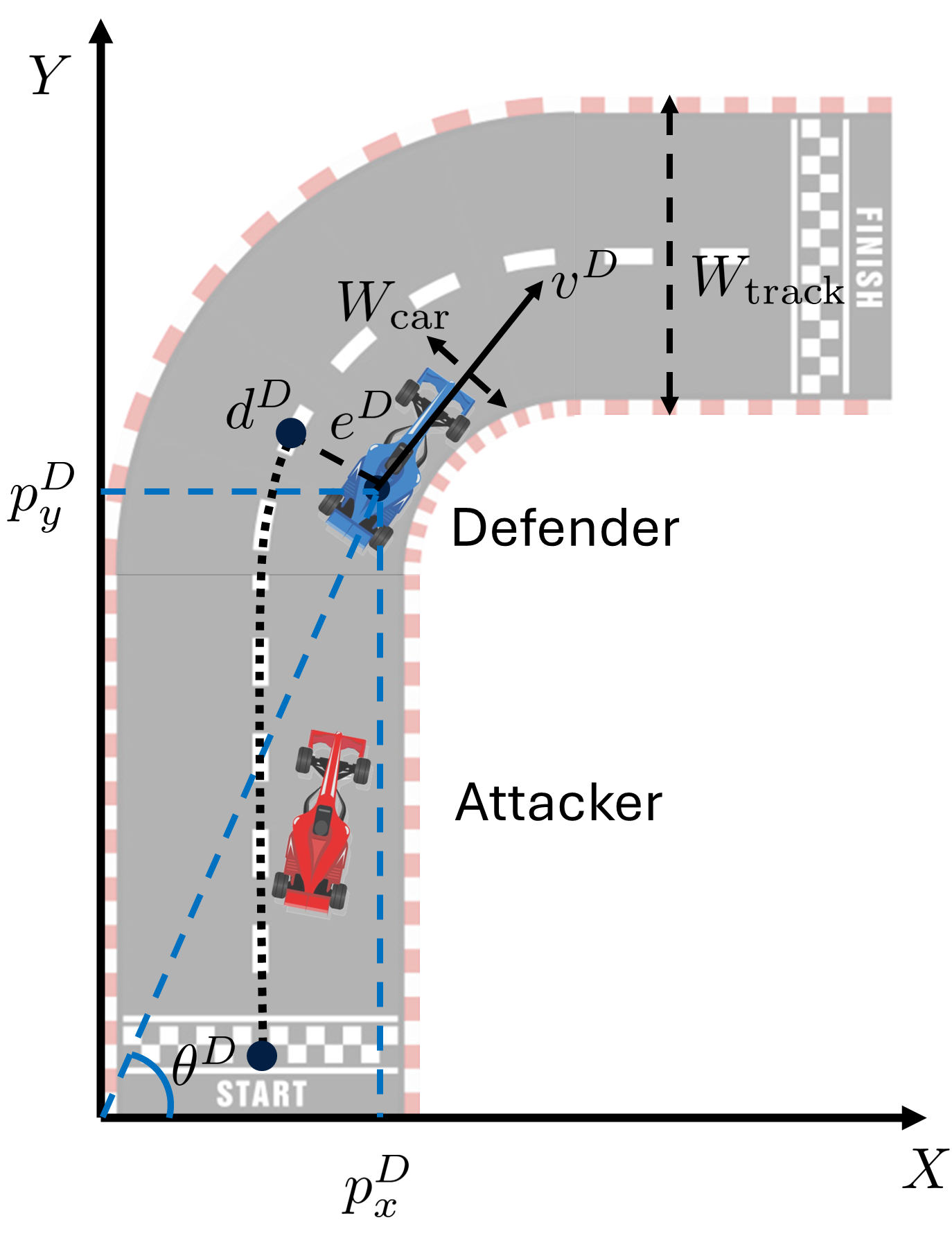}
    % \caption{\textcolor{blue}{Visual representation of roles and states of racing vehicles, and parameters of the racing track.}}
        \caption{Visual representation and pertinent parameters in a car racing scenario.}
    \label{Fig: kinetic model}
\end{figure}

In this paper, we consider a racing game between two vehicles. Specifically, we consider the leading vehicle as the defender (D) and the following vehicle as the attacker (A). The set of racing vehicles is then defined as $\mathcal{N}=\{A,D\}$. For simplicity, we assume that both racing vehicles possess the same kinematics. For vehicle $i\in\mathcal{N}$, we consider the following state vector $x^i_\tau=[p^i_{x,\tau},p^i_{y,\tau},\theta^i_{\tau},v^i_{\tau}]$, where $\tau$ is the time stamp such that $\tau\in\mathcal{T}$, $\mathcal{T}=\{0,1,...,T\}$, $T$ is the maximum time stamp. As in Fig.~\ref{Fig: kinetic model}, $p^i_{x,\tau}$ is the global-X coordinates, $p^i_{y,\tau}$ is the global-Y coordinates, $\theta^i_\tau$ is the heading angle, and $v^i_{\tau}$ is the longitudinal velocity, all with respect to time stamp $\tau$. In addition, the vehicle position can be transformed into the Frenet–Serret Coordinate $(d^i_\tau, e^i_\tau)$, where $d^i_\tau$ denotes the vehicle's traveling distance along the track centerline and $e^i_\tau$ denotes the vehicle's lateral distance to the centerline, while the total track width is $W_{\text{track}}$ and vehicle width is $W_{\text{car}}$. Moreover, the input vector at $\tau$ is defined as $r^i_\tau=[\alpha^i_\tau,\delta^i_\tau]$, where $\alpha^i_\tau$ is the acceleration and $\delta^i_\tau$ is the steering angle. The discrete vehicle kinematics, $x^i_{\tau+1}=f(x^i_\tau,r^i_\tau)$, is then characterized by the following equations~\cite{tassa2014control}:
\begin{equation}
\label{dynamics}
\left\{
\begin{aligned}
p^i_{x,\tau+1} &= p^i_{x,\tau}+f_r(v^i_\tau,\delta^i_\tau)\cos(\theta^i_\tau),\\
p^i_{y,\tau+1} &= p^i_{y,\tau}+f_r(v^i_\tau,\delta^i_\tau)\sin(\theta^i_\tau),\\
\theta^i_{\tau+1} &= \theta^i_{\tau}+\arcsin(\frac{\tau_s v^i_\tau\sin(\delta^i_\tau)}{b}),\\
v^i_{\tau+1} &= v^i_\tau+\tau_s\alpha^i_\tau,
\end{aligned}
\right.
\end{equation}
where $\tau_s$ is the time interval and $b$ is the wheelbase. The function $f_r(v,\delta)$ is defined as
\begin{equation}
f_r(v,\delta) = b+\tau_sv\cos(\delta)-\sqrt{b^2-(\tau_sv\sin(\delta))^2}.
\end{equation}

In addition to the kinematics, we impose constraints on maximum velocities of racing vehicles, such that 
\begin{equation}
    v^i_\tau\leq v^i_\textup{max}, \forall i \in \mathcal{N}.
    \label{maxv}
\end{equation}
Note that the maximum velocity of a racing vehicle is subject to multiple factors such as air resistance and tire condition. These factors may vary from time to time and from vehicle to vehicle. Therefore, the maximum velocities of different racing vehicles can be different.

% We formulate the autonomous racing versus game as a multi-agent trajectory planning problem. 
% %
% In the racing game, two vehicles drive on a closed racing track. According to their initial position, we classify them as the leader (L) and follower (F). On the 2D plane, the vehicle states are x-y position and orientation as $\mathcal{S}=(x, y, \theta)$. The actions are vehicle velocity and steering wheel angle as $\mathcal{A} = (V, \delta)$. The vehicle follows the bicycle kinetic model, 
% %
% \begin{equation}
% \begin{bmatrix}
% \dot{x} \\
% \dot{y} \\
% \dot{\theta}
% \end{bmatrix}
% =
% \begin{bmatrix}
% V \cos(\theta) \\
% V \sin(\theta) \\
% \frac{V \tan(\delta)}{L}
% \end{bmatrix}
% \end{equation}
% %
% At the time $t$, the overall states and actions are the contamination of two cars as $s_t=\{s_t^L, s_t^F\}$ and $a_t=\{a^L_t, a^F_t\}$. While the relative position of two cars may change during the racing, we keep their name of leader and follower based on their initial condition. 

% We define the racing track in a 2D x-y space. For simplicity, the racing track has the same width and the track edges are bounded by no-passing barriers. During the game, two cars are placed around the starting line position with close initial velocity. Their target is to pass the finish line with minimal time. Therefore, the versus racing game is formulated as a trajectory planning problem,

% \begin{equation}
% \begin{aligned}
    
% \end{aligned}
% \end{equation}

% It has start and ending lines 

\subsection{Monte Carlo Tree Search}
MCTS is a heuristic search algorithm that has been successfully applied to search for near-optimal strategies in various multi-agent games, including board games~\cite{silver2017mastering}, multi-robot planning~\cite{riviere2021neural}, and autonomous driving~\cite{li2022efficient}. Vanilla MCTS algorithm maintain a subtree of the original game tree, and performs iterations to approximate the optimal gaming strategy for all players. In each iteration, four steps are performed in a sequential manner.
\begin{itemize}
    \item \textbf{Selection}. From the root node of the game tree, the algorithm recursively selects a child node until a leaf is reached.
    \item \textbf{Expansion}. An unexplored action of the current leave node is selected, and the new node reached through performing this action is appended to the game tree.
    \item \textbf{Simulation}. The game is played to the end from the newly added node by selecting actions of players following some roll-out policy.
    \item \textbf{Back-propagation}. Based on the simulation outcomes, values of the nodes along the path are updated.
\end{itemize}

The algorithm terminates after a given number of iterations is reached. At each node of the game tree, the optimal strategy of the player is determined as performing the action selected the most through the selection step.
\section{Method} \label{Sec: method}

% In this section, we introduce the methods of trajectory planning with SPS (SPS) rules. We first formulate the trajectory planning problem (Sec.~\ref{Subsec: problem formulation}) and define the utilities for versus game and one-motion and enough-space SPS rules as  (Sec.~\ref{Subsec: sps define}). Then we propose the two-level planning method with game formulation (Sec.~\ref{Subsec: planning with game}).

In this Section, we introduce the method of trajectory planning with SPS rules. We first define the one-motion and enough-space SPS rules (Sec.~\ref{Subsec: sps define}). Then we propose the two-level planning method with game formulation (Sec.~\ref{Subsec: planning with game}) and the corresponding solving scheme.

\begin{figure*}
    \centering
    \includegraphics[width=0.9\linewidth]{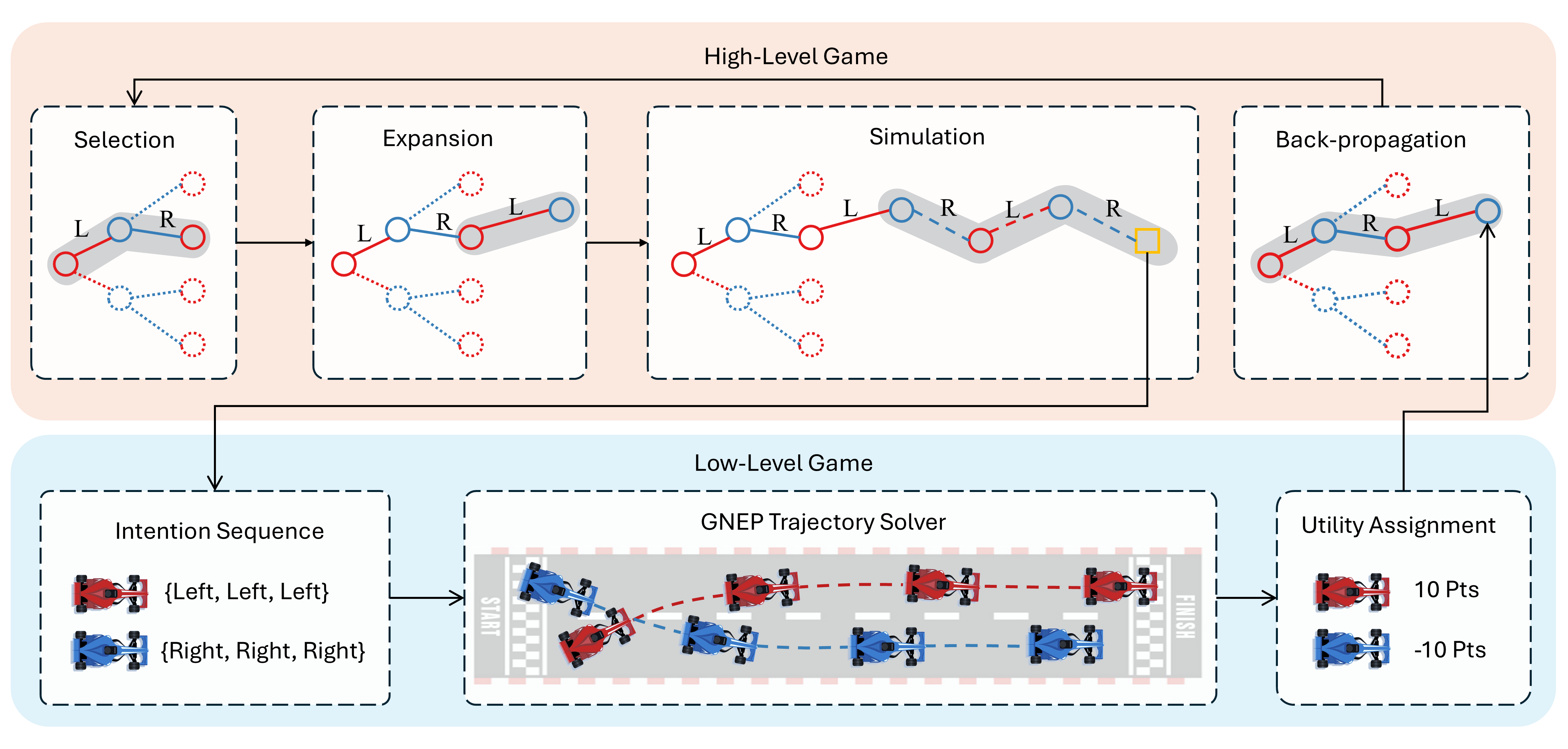}
    \caption{Pipeline of the proposed algorithm. In the high-level game, the MCTS algorithm is applied to search for optimal strategies with respect to racing intention. In the low-level game, GNE trajectories are obtained conditioned on high-level intentions, which are obtained through forward propagation of high-level MCTS. Utilities are then determined and provided back to the high-level game solver for back-propagation.}
    \label{Fig: pipeline}
\end{figure*}

% \begin{figure}
%     \centering
%     \includegraphics[width=0.8\columnwidth]{figures/Illustration1.png}
%     \caption{SPS illustration}
%     \label{Fig: sps illustration}
% \end{figure}

% \subsection{Planning problem formulation} \label{Subsec: problem formulation}

% The versus racing game is a trajectory optimization problem for vehicle $i$. The trajectory is a sequence of state and actions from $\tau=0$ to $T$ as $\{(x^t_\tau, u^i_\tau)\}_{\tau=0,1,\dots, T}$. $i=\{D, A\}$ represents defender and attacker. The optimization problem is formulated as,
% \begin{align}
% & \quad \max_{x^i_\tau, u^i_\tau} \sum_{\tau=0}^{T-1} r_\tau^i(x^i_\tau, u^i_\tau) + r^i_T(x^i_T, u^i_T) \\
% & \text{s.t. } x^i_{\tau+1} = f(x^i_\tau, u^i_\tau), \forall \tau \in [0, T-1] \label{Eqn: opt-dynamic} \\
% & \qquad x^i_\tau \in [x^i_{\tau,\text{low}}, x^i_{\tau,\text{up}}] \label{Eqn: opt-x-inequ} \\
% & \qquad u^i_\tau \in [u^i_{\text{low}}, u^i_{\text{up}}] \label{Eqn: opt-u-inequ}
% \end{align}
% %
% where $r_\tau^i(x^i_\tau, u^i_\tau)$ and $r^i_T(x^i_T, u^i_T)$ are procedure and terminal rewards for each vehicle. Eqn.~\eqref{Eqn: opt-dynamic} represents the vehicle kinematic model and Eqn.~\eqref{Eqn: opt-x-inequ} and \eqref{Eqn: opt-u-inequ} are state and action limits defined by track boundary and vehicle capability. 
% The target of optimization is to find an equilibrium that both the attacker and the defender maximize their own rewards to ensure victory and safety. Especially, their rewards functions are different and we detailed the design of rewards for racing and SPS in the following section.

\subsection{SPS Model} \label{Subsec: sps define}

% In racing vehicle competitions, the drivers maximize the velocity to reach the finish line with minimal time. Meanwhile, they must also avoid collision with track boundaries and obey SPS to ensure fair and safe competition. In professional games like Formula One, the judge and steward will estimate the driving behavior of each vehicle to penalize unsportsmanlike maneuvers. While SPS are wide-ranging and diverse across different games, in this paper, we focus on two mostly common SPS: the one-motion rule and the enough-space rule. In the following, we formulate the utility function for the racing competition and define concrete criteria to judge the violation of two SPS.
% %
% \begin{equation} \label{Eqn: drive reward}
%     r^i_{\text{drive}}(x^i_\tau, a^i_\tau) = w_{d1} v^i_\tau - w_{d2}\|\delta^i_\tau\| - w_{d2} C(p^i_{x,t}, p^i_{y,t})
% \end{equation}
% %
% Eqn.\eqref{Eqn: drive reward} shows the rewards for racing. The first term encourages the vehicle to accelerate to achieve higher speed. The second term regularizes the steering to reduce abrupt turning. And the third term $C$ penalizes the collision when the vehicle crashes the track boundary or other cars. $w_{d1}, w_{d2}, w_{d3}$ are weights.
% %
% \begin{equation} \label{Eqn: overtake reward}
%     r^i_{\text{overtake}}(x^d_T, x^a_T) = d^D_T - d^A_T
% \end{equation}
% %
% Eqn.~\eqref{Eqn: overtake reward} denotes the terminal distance that the defender is in ahead of the attacker. This reward is a crucial criterion to judge the victory of the game. 

To estimate the violation of SPS rules, we first define the binary ``block'' function. The binary ``block'' function is defined as $\text{Block}(x^D_\tau, x^A_\tau) = (d^D_\tau > d^A_\tau) \cap |e^D_\tau - e^A_\tau| \leq W_{\text{car}} \in \{0, 1\}$, which means the defender is in front of the attacker and two cars are laterally overlapping each other. In this case, we think the defender is blocked. Then we can define two SPS rules, both of which aim to constrain the gaming behavior of the defender.

\begin{definition}
    \textbf{One-motion rule}: Given a time span, the defender violates the one-motion rule if there exists 4 time steps $\tau_1 < \tau_2 < \tau_3 < \tau_4$ such that
\begin{equation}
\begin{aligned}
    \hat{F}_{\text{OM}}(x^D_{\tau_{1:4}}, x^A_{\tau_{1:4}}) = & \neg \text{Block}(x^D_{\tau_1}, x^A_{\tau_1}) 
     \cap \text{Block}(x^D_{\tau_2}, x^A_{\tau_2}) \\ &
     \cap \neg \text{Block}(x^D_{\tau_3}, x^A_{\tau_3}) \cap \text{Block}(x^D_{\tau_4}, x^A_{\tau_4})
\end{aligned}    
\end{equation}
is equal to 1. $\neg$ and $\cap$ denote negative and `and' in logic operations. We denote the trajectory of player $i$ as $X^i:=\{x^i_\tau\}_{\tau\in\mathcal{T}}$, and the one-motion rule is defined as
\begin{equation}
\begin{aligned} \label{Eqn: SPS one-motion}
    &F_{\text{OM}}(X^D, X^A) = \cup_{\tau_1 < \tau_2 < \tau_3 < \tau_4} (\hat{F}_{\text{OM}}(x^D_{\tau_{1:4}}, x^A_{\tau_{1:4}})).
\end{aligned}    
\end{equation}
$F_{\text{OM}} \in \{0,1 \}$ is the binary flag to represent the violation of one-motion rule. Intuitively, it means if the defender take lateral movements to block twice, then it violates the one-motion SPS. 
\end{definition}

\begin{definition}
    \textbf{Enough-space rule}: Given a time period, the defender violates the enough-space rule, if there exists 2 time steps, $\tau_1 < \tau_2$, such that
\begin{equation}
\begin{aligned}
    \hat{F}_{\text{ES}}(x^D_{\tau_{1,2}}, &x^A_{\tau_{1,2}}) =\neg \text{Block}(x^D_{\tau_1}, x^A_{\tau_1}) \cap (v^A_{\tau_1} - v^D_{\tau_1} > \Delta v) \\
    &\cap(0.5 W_{\text{track}} - | e^A_{\tau_1} | \leq W_{\text{car}}) \cap \text{Block}(x^D_{\tau_2}, x^A_{\tau_2})
\end{aligned}
\end{equation}
is equal to 1, where $\Delta v$ is the velocity threshold. The enough-space rule is defined as
\begin{equation} \label{Eqn: SPS enough space}
    F_{\text{ES}}(X^D, X^A) = \cup_{\tau_1 < \tau_2}(\hat{F}_{\text{ES}}(x^D_{\tau_{1,2}}, x^A_{\tau_{1,2}})).
\end{equation}
Intuitively, when the attacker has higher speed and try to overtake near the track boundary, the blocking movement of the defender is regarded as the violation of enough-space SPS.
\end{definition}

% the F is flag, binary value {0,1}.
% need two figures to illustrate these two ideas.

% Finally, we can integrate all rewards for two cars. During their driving, they share the same reward function as ,
% \begin{equation}
    % r^D_\tau = r^A_\tau = w_0 r^i_{\text{drive}} - w_1 F^i_{\text{OM}} - w_2 F^i_{\text{ES}}
% \end{equation}

% \begin{equation}
    % r^D_T = w_3 r^i_{\text{overtake}}; \quad r^A_T = -w_4 r^i_{\text{overtake}}
% \end{equation}

\subsection{Planning with Two-Level Game Formulation} \label{Subsec: planning with game}

From Section \ref{Subsec: sps define}, it is clear that the SPS rules are non-differentiable. In this Section, we introduce a hierarchical game formulation to incorporate SPS rules in the racing game.
At the higher level, we focus on the gaming behavior over the racing intentions of the competitors, which are usually low-frequency, discrete in nature, and coupled with each other. These racing intentions are usually strongly relevant to the SPS rules. As a result, the optimal gaming strategies for both competitors are obtained with compliance of SPS rules.
At the lower level, we generate the corresponding GNE trajectories for both competitors, which are conditioned on high-level strategies. By obtaining the trajectories, the utilities of competitors can be determined, which are then fed back into the higher-level game solver to determine the optimal strategy.
Through two levels of game formulation, the requirements of the SPS model are guaranteed to be satisfied, and the optimal strategies and trajectories are obtained simultaneously. The pipeline of the proposed algorithm is shown in Fig. \ref{Fig: pipeline}.

\textbf{High-level Game}. Similar to~\cite{li2022efficient}, we model the intention-level game of the players as an extensive-form Stackelberg game. In this game, rounds of playing are involved, where players act in turns in each round until the maximum number of rounds $M$ is reached. We define the action set as $\mathcal{A}$, and the set of history as $\mathcal{H}$. Specifically, a history $h\in\mathcal{H}$ is defined as the sequence of past actions taken by players: $h=[a^1_1,a^2_1,...]$, where $a^i_\tau$ is the action of $i$ at round $t$. It is then obvious that each node at the game tree is associated with a unique $h$. For a player $i\in\mathcal{N}$, its policy is defined as a map from the history set to the action set $\pi^i:\mathcal{H}\rightarrow \mathcal{A}$, such that for a history $h\in\mathcal{H}$, $a=\pi^i(h)\in\mathcal{A}$ denotes an action took by $i$ given this history. For player $i$, the value function given all other players' policy $\pi^{-i}$, is given as
\begin{equation}
\label{value}
    V^i_{\pi^{-i}}(h)=\left\{
    \begin{aligned}
    &\max_a V^i_{\pi^{-i}}(h\oplus a)\ &i\ \textup {to play}\\
    &V^i_{\pi^{-i}}(h\oplus\pi^{-i}(h))\ &-i\ \textup{to play}\\
    &u^i(h)\ &h \textup{ is end of game}
    \end{aligned},
    \right.
\end{equation}
where in the above equation, $h\oplus a$ denotes the new history attained by appending action $a$ to the current history $h$, and $u^i(h)$ denotes the utility of player $i$ at the end of the game with history $h$. With the definition of the value function, we formally introduce the Nash equilibrium of the game.
\begin{definition}
A Nash equilibrium of the game is a set of policies $\{\pi^{i*}\}_{i\in\mathcal{N}}$, such that 
$\pi^{i*}(h)=\arg\max_a V^i_{\pi^{-i*}}(h\oplus a)$ for all $i\in\mathcal{N}$ and for all $h\in\mathcal{H}$.
\end{definition}

The above definition implies that when a Nash equilibrium is reached, the policy of each player is optimal at each state given the equilibrium policies of all rivals. We set the ending criterion of the game as when the maximum number of playing rounds is reached. The action set $\mathcal{A}$ is defined as the set of lateral strategies of the racing car, which is featured by a set of discretized lateral displacements. Vanilla MCTS algorithm is used to solve for the Nash equilibrium approximately. Specifically, at a node with history $h$ and player $i$ to play, the UCT~\cite{kocsis2006bandit} formula is used for selection:
\begin{equation}
    \underset{a\in\mathcal{A}}{\arg\max} \frac{Q^i(h,a)}{N(h\oplus a)} + c\sqrt{\frac{\textup{ln}(N(h))}{N(h\oplus a)}},
\end{equation}
while $Q^i(h,a)$ is the total utility of playing $a$ at $h$ for player $i$. $N(h)$ is the total number of simulations performed in which $h$ is selected. $c$ is a predefined constant. The selection process is performed recurrently until a leaf node at the current sub-tree is reached, where a leaf node is a node with unexplored actions. Then, an unexplored action is selected, which results in a new node appended to the tree. From the newly appended node onward, a random roll-out policy is applied until the end of the game, and the utility $u^i$ is obtained for all $i\in\mathcal{N}$. In the back-propagation, the value of all the nodes along the path traversed in the selection and expansion step are updated through the following formula:
\begin{equation}
\begin{aligned}
    Q^i(h,a)&\leftarrow Q^i(h,a) + u^i,\\
    N(h)&\leftarrow  N(h) + 1.
\end{aligned}
\end{equation}
Eventually, the optimal strategy is given by 
\begin{equation}
    \pi^{i*}(h)=\underset{a}{\arg\max}N(h\oplus a).
\end{equation}

\textbf{Low-level Game}. In the previous formulation, one problem unsolved is how to determine the utility function $u^i(h)$ for $h$ and $i$ when $h$ is a history at the end of the game. In our study, utilities corresponding to a player at the end of the game depend on two factors: 1) whether the player is taking the lead, and 2) whether its behavior violates the requirements of the SPS model. These two factors can only be determined when the actual trajectories of all players are obtained. At the lower-level game formulation, we formulate the interaction of players as a GNEP conditioned on $h$. In this problem, each player follows a feasible trajectory that minimizes its own cost conditioned on $h$, while trajectories of all players are required to satisfy joint constraints such as collision avoidance constraints. The GNEP is formulated as follow:

\begin{equation}
\label{problem}
\begin{aligned}
\min_{X^i}\ &C^i(X^i,X^{-i}|h)\\
\textup{s.t.}\ &G(X^i)\leq 0,\\
&H(X^i,X^{-i})\leq 0
\end{aligned}
\end{equation}
where $X^{-i}$ denotes the trajectories of all players except $i$. $C^i$ is the cost function of player $i$, which is conditioned on $h$. Specifically, the cost function $C^i$ aims to measure two aspects of the trajectory $X^i$, namely 1) how well the trajectory $X^i$ matches the lateral strategy described by $h$, and 2) the longitudinal progress of $X^i$. Note that it does not measure whether the requirement of the SPS model is satisfied, as this is handled at the high-level game formulation. $G$ enforces the kinematic constraint (\ref{dynamics}) and the velocity constraint (\ref{maxv}). $H$ is the coupled collision avoidance constraint enforced over all trajectories to maintain a safety distance $d_\textup{safe}$ between all players. Problem (\ref{problem}) is an optimization problem for all players $i\in\mathcal{N}$. Solving for problem (\ref{problem}) results in a GNE, which is formally defined as follows:
\begin{definition}
    A set of trajectories, $X^*=\{X^{i*}\}_{i\in\mathcal{N}}$, is said to be a GNE to problem (\ref{problem}), if it satisfies $H(X^*)\leq0$ and $G(X^{i*})\leq 0\ \forall i$, and $X^{i*}$ is a minimizer to the optimization problem $C^i(X^i,X^{-i*}|h)$ over $X^i$ for all $i$.
\end{definition}

To solve problem (\ref{problem}) and obtain the corresponding GNE $X^*$, we use the standard Iterative Best Response (IBR) method. At each iteration, we fix the current solution of $X^{-i}$ and solve for the optimal solution $X^i$ of problem (\ref{problem}), and we iterate the process until all $X^i$ converge to $X^{i*}$. Specifically, at each iteration, the following optimization problem is to be solved:
\begin{equation}
\begin{aligned}
\quad \max_{x^i_\tau, r^i_\tau}\ &\sum_{\tau=0}^{T} C_\tau^i(x^i_\tau, r^i_\tau|h)\\
\text{s.t. }\ &x^i_{\tau+1} = f(x^i_\tau, r^i_\tau), \\ &v^i_\tau<v^i_\textup{max},\\
&d(x^i_\tau,x^{-i}_\tau)>d_\textup{safe},\forall \tau \in \mathcal{T}.
\end{aligned}
\label{game1}
\end{equation}
$d(x^i_\tau,x^{-i}_\tau)$ measures the Euclidean distance between the center of $i$ and $-i$ at $\tau$. The cost function $C^i_\tau$ is defined as
\begin{equation}
C_\tau^i(x^i_\tau, r^i_\tau|h) = w_1 (d_\tau(x^i_\tau|h))^2+(v^i_\tau-v^i_\textup{max})^2.
\end{equation}
$w_1$ is predefined coefficients, and $d_\tau$ measures the lateral distance between $x^i_\tau$ and the lateral strategy defined by $h$ at time $\tau$. 

Noted that problem (\ref{game1}) is a nonlinear optimization problem, solving (\ref{game1}) directly is time-consuming and may suffer from sub-optimality. Therefore, we propose to solve it approximately with a two-stage solver. Consider the simplified state vector $\hat{x}^i_\tau=[p^i_{x,\tau},p^i_{y,\tau}]$, and further, suppose that the tangent angle of the racing track at $\hat{x}^i_\tau$ is $\xi(\hat{x}^i_\tau)$, we approximate problem (\ref{game1}) with the following problem:
\begin{equation}
\begin{aligned}
\quad \max_{\hat{x}^i_\tau}\ &\sum_{\tau=0}^{T-1} \hat{C}_\tau^i(\hat{x}^i_\tau,\hat{x}^i_{\tau+1}|h)\\
\text{s.t. }\ &d(\hat{x}^i_\tau,\hat{x}^i_{\tau+1})<v^i_\textup{max}\tau_s, \\ 
&|\textup{Arg}(\hat{x}^i_{\tau+1}-\hat{x}^i_\tau)-\xi(\frac{\hat{x}^i_\tau+\hat{x}^i_{\tau+1}}{2})|\leq\gamma_\textup{max},\\
&d(\hat{x}^i_\tau,\hat{x}^{-i}_\tau)>d_\textup{safe},\forall \tau \in \mathcal{T}.
\end{aligned}
\label{game2}
\end{equation}
$\textup{Arg}(\cdot)$ returns the heading angle of the vector. In problem (\ref{game2}), the first constraint ensures that the velocity does not exceed the limit. The second constraint ensures that the heading angle of the racing car does not deviate from the tangent angle of the racing track, and therefore the trajectory is smooth.
Further, $\hat{C}_\tau^i$ is given as
\begin{equation}
\begin{aligned}\hat{C}_\tau^i(\hat{x}^i_\tau,\hat{x}^i_{\tau+1}|h) &= w_1 (d_\tau(\hat{x}^i_\tau|h))^2\\
&+(d(\hat{x}^i_\tau,\hat{x}^i_{\tau+1})/\tau_s-v^i_\textup{max})^2.
\end{aligned}
\end{equation}
Problem (\ref{game2}) can be solved by discretizing the racing space and dynamic programming. GPU parallelization is applied to accelerate the computation. During IBR, we solve problem (\ref{game2}) instead of problem (\ref{game1}) until convergence, and then problem (\ref{game1}) is solved by tracking the generated trajectory to ensure kinematic feasibility. 
Since the obtained trajectories are conditioned on $h$, the set of GNE trajectories can be viewed as a function of $h$. The utility function in equation (\ref{value}) can then be reformulated as $u^i(X^*(h))$. We then determine the value of $u^i$ by measuring the leading distance of player $i$ at the end of the trajectory $X^{i*}$. Meanwhile, if $X^{i*}$ violates the requirements of SPS, a large penalty is imposed on $u^i$. Through the low-level game, the utilities for all players are determined, which is then used by the high-level game solver for back-propagation.

Specifically, for a game involving a defender and an attacker, the utilities are given as
\begin{equation}
\begin{aligned}
    u^A(h) &= \beta\textup{prog}(X^{A*}(h))-\textup{prog}(X^{D*}(h))-\textup{Reg}(h),\\
    u^D(h) &= \beta\textup{prog}(X^{D*}(h))-\textup{prog}(X^{A*}(h))\\
    &-\omega \textup{SPS}(X^{D*}(h),X^{A*}(h)),
\end{aligned}
\end{equation}
where $\textup{prog}(\cdot)$ measures the total progress along the track. $\omega$ and $\beta$ are predefined constants. $\textup{Reg}(\cdot)$  is a small regularization term to penalize inconsistent strategies of the attacker. This is added mainly to eliminate randomness in strategies of the attacker. $\textup{SPS}(X^D,X^A)$ is a binary function, which is equal to 1 if $x^D$ violates the SPS given $X^A$ and 0 otherwise. $\textup{SPS}\in\{F_\textup{OM},F_\textup{ES},F_\textup{OM}\cup F_\textup{ES}\}$
% \begin{equation}
%     \textup{SPS}(X^D|X^A) = \cup_{\tau=1}^T \left (F_{\text{OM}}(x^D_{\tau}, x^A_{\tau}) \cup F_{\text{ES}}(x^D_{\tau}, x^A_{tau}) \right)
% \end{equation}
where $F_{\text{OM}}$ and $F_{\text{ES}}$ are one-motion rule and enough space rule defined in Eqn.~\eqref{Eqn: SPS one-motion} and \eqref{Eqn: SPS enough space}.

% \begin{figure}
%     \centering
%     \includegraphics[width=0.8\columnwidth]{figures/Illustration1.png}
%     \caption{illustration of method}
%     \label{Fig: illustration2}
% \end{figure}
\section{Experiment} \label{Sec: experiment}

\subsection{Experiment Settings}

To thoroughly examine the effect of SPS, we consider the following three cases:

1) Only SPS1 is effective.

2) Only SPS2 is effective.

3) Both SPS1 and SPS2 are effective.

To reveal the impact of SPS on the gaming behaviors of the players and the final outcomes, in each case, we inspect the gaming behaviors of racing agents with the following settings:

1) Both players do not know the existence of SPS.

2) Both players know the existence of SPS.

3) The attacker knows the existence of SPS, but the defender does not know. 

4) The defender knows the existence of SPS, but the attacker does not know. 

We assume that under the above settings, both players perform a rational strategy. Settings 1 and 2 can be easily realized by enabling/disabling penalties corresponding to SPS in the utility function $u^i$ for all $i$. To implement settings 3 and 4, we perform a two-stage solving scheme. Suppose that player $i$ is the attacker. Under setting 3, we first solve the game by enabling penalties for SPS. As a result, the rational strategy of player $i$ with the knowledge of SPS is obtained. At the second stage, we fix the strategy of player $i$ and solve for the strategy of player $-i$, which is the defender, with the SPS penalties disabled. Therefore, the defender will perform the optimal strategy with respect to the strategy of player $i$, under a situation where SPS does not exist. We do the opposite under setting 4, while penalties for SPS are first disabled while solving for the strategy of player $i$, and then enabled while solving for the strategy of player $-i$.

In the high-level game, the action space of each player is set to be $\{-1\,\textup{m},1\,\textup{m}\}$, which is the intended lateral displacement with respect to the centerline of the track. The planning horizon is set to be 6\,s. Other parameters are shown in Table I.

\begin{table}[]
\caption{Parameters used in experiments}
\centering
\begin{tabular}{cc|cc|cc}
\hline
Param.            & Value  & Param.   & Value & Param.  & Value \\ \hline
c                 & 30     & $\omega$ & 15    & $\beta$ & 1.1   \\
$d_\textup{safe}$ & 1.8\,m & M        & 3     & T       & 15    \\
$\tau_s$          & 0.4\,s & $\xi_\textup{max}$ & 0.16     & $w_1$     &  100     \\ 
$W_\textup{track}$   & 5.8\,m & $W_\textup{car}$ & 1.8\,m     & $\Delta v$     &  1.5\,m/s     \\ \hline
\end{tabular}
\end{table}

\begin{figure*}[t]
\centering
\subfigure[Setting 1 on the straightway with SPS1]{\includegraphics[scale=0.43]{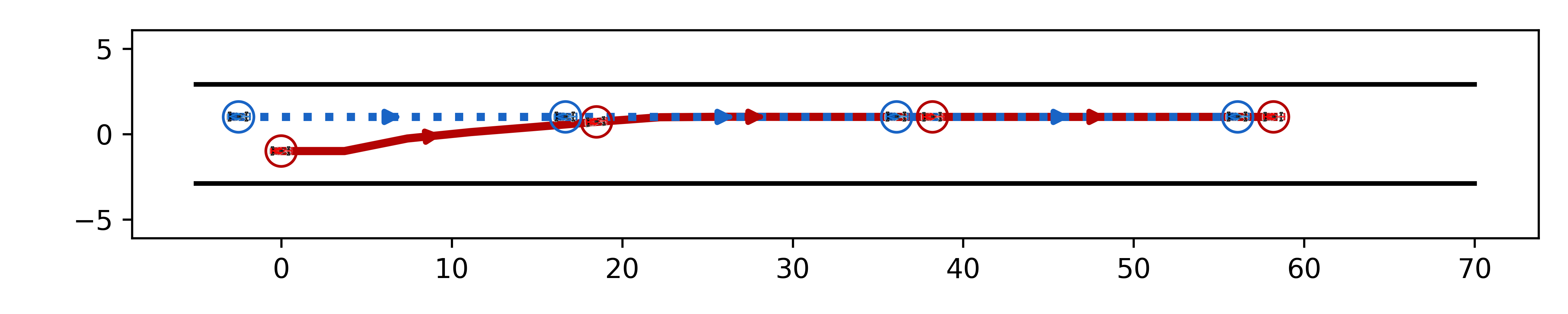}}
\subfigure[Setting 2 on the straightway with SPS1]{\includegraphics[scale=0.43]{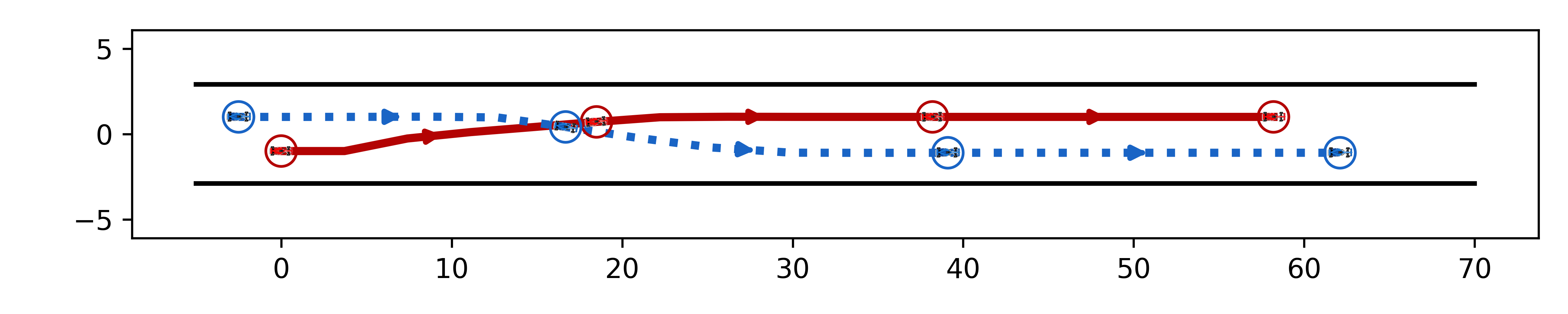}}
\subfigure[Setting 3 on the straightway with SPS1]{\includegraphics[scale=0.43]{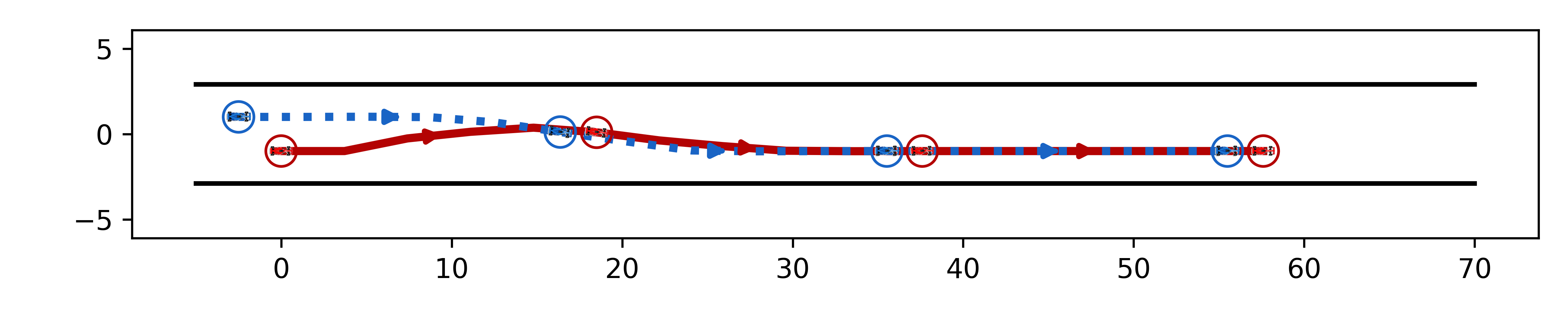}}
\subfigure[Setting 4 on the straightway with SPS1]{\includegraphics[scale=0.43]{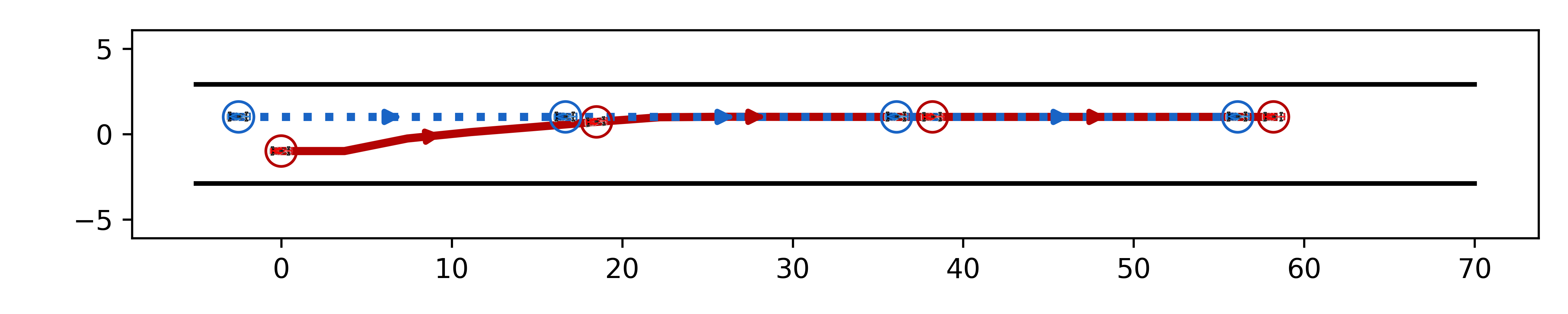}}
% \subfigure[Setting 1]{\includegraphics[width=0.47\textwidth]{figures/main_straight/1.png}} \hfill
% \subfigure[Setting 2]{\includegraphics[width=0.47\textwidth]{figures/main_straight/2.png}} 
% \vspace{2pt}
% \subfigure[Setting 3]{\includegraphics[width=0.47\textwidth]{figures/main_straight/3.png}} \hfill
% \subfigure[Setting 4]{\includegraphics[width=0.47\textwidth]{figures/main_straight/4.png}}
\caption{Results on the straightaway under SPS1. In (a) where none of the players know the existence of SPS1, the defender (racing car in red) switches to the front of the defender (racing car in blue) to perform a blocking. Since the attacker does not know the existence of SPS1, it considers overtaking to be impossible and thus decides to follow the attacker. In (b) where both players know SPS1, the defender also performs blocking to delay the overtaking, while the attacker decides to switch to the lower half of the track. Due to SPS1, the defender knows that a second blocking is forbidden, so it stays in the upper half of the track to let the attacker pass. In (c) where only the attacker knows the existence of SPS1, it attempts to perform an overtaking by switching to the lower half of the track after the blocking, similar to the case in (b). However, since the defender does not know SPS1, it continues to block the attacker, and thus the attacker fails to overtake. In (d), although the defender knows SPS1, the attacker does not, so it cannot come up with the correct strategy as in (b). Therefore, it also fails to perform an overtaking.}
\label{fig:strateghtSPS1}
\end{figure*}

\begin{figure*}[t]
\centering
\subfigure[Setting 1 at corner with SPS1]{\includegraphics[scale=0.28]{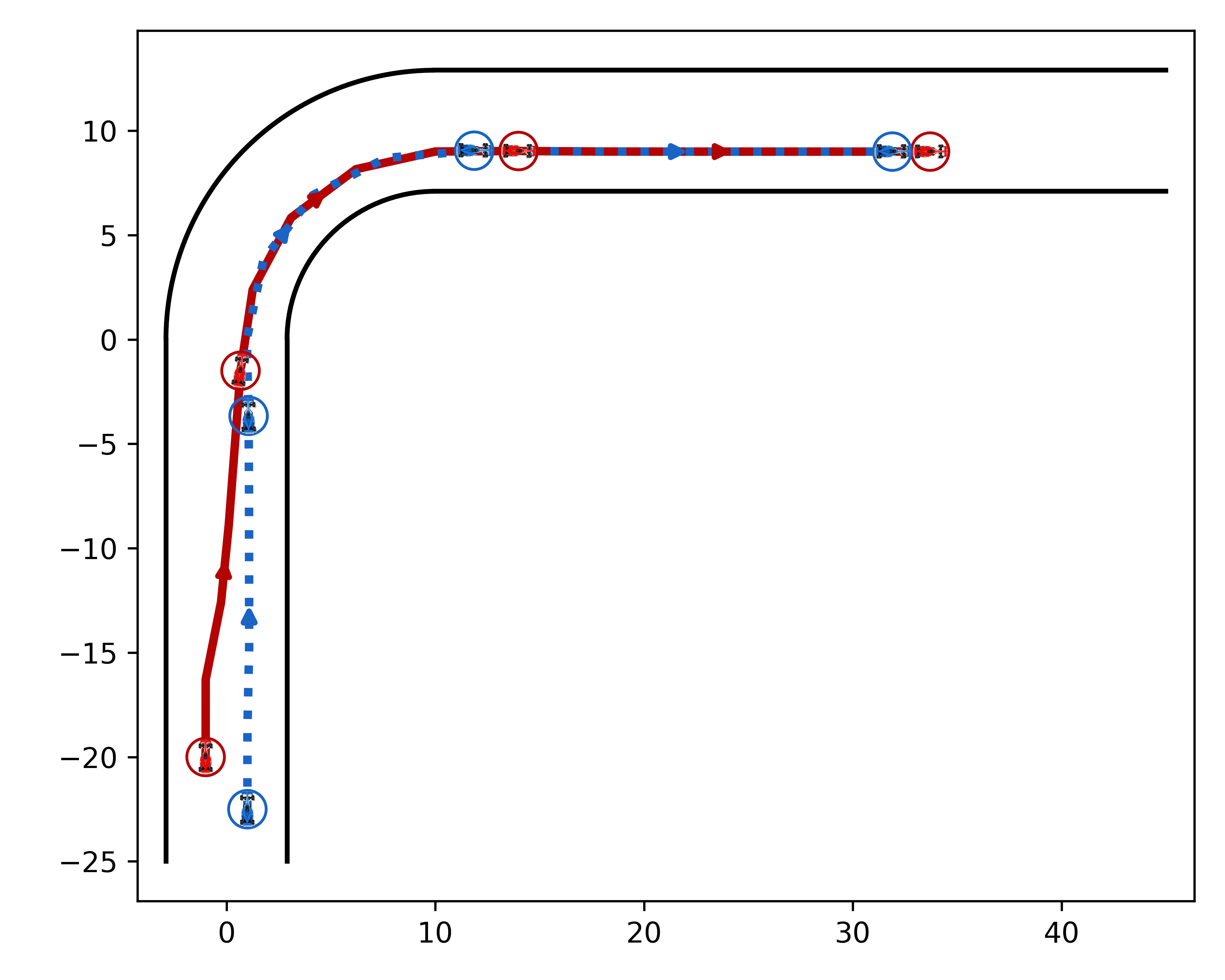}}
\subfigure[Setting 2 at corner with SPS1]{\includegraphics[scale=0.28]{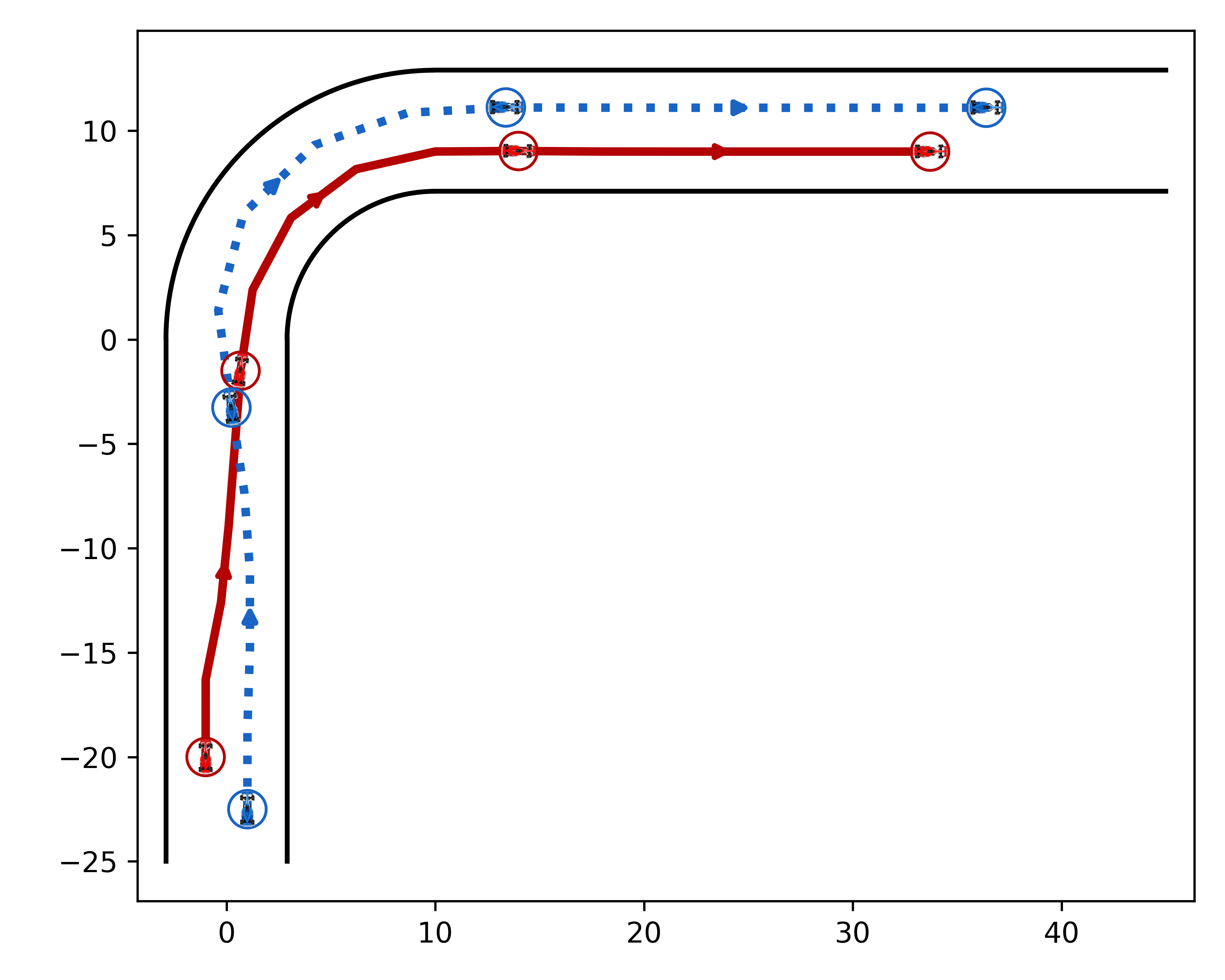}}
\subfigure[Setting 3 at corner with SPS1]{\includegraphics[scale=0.28]{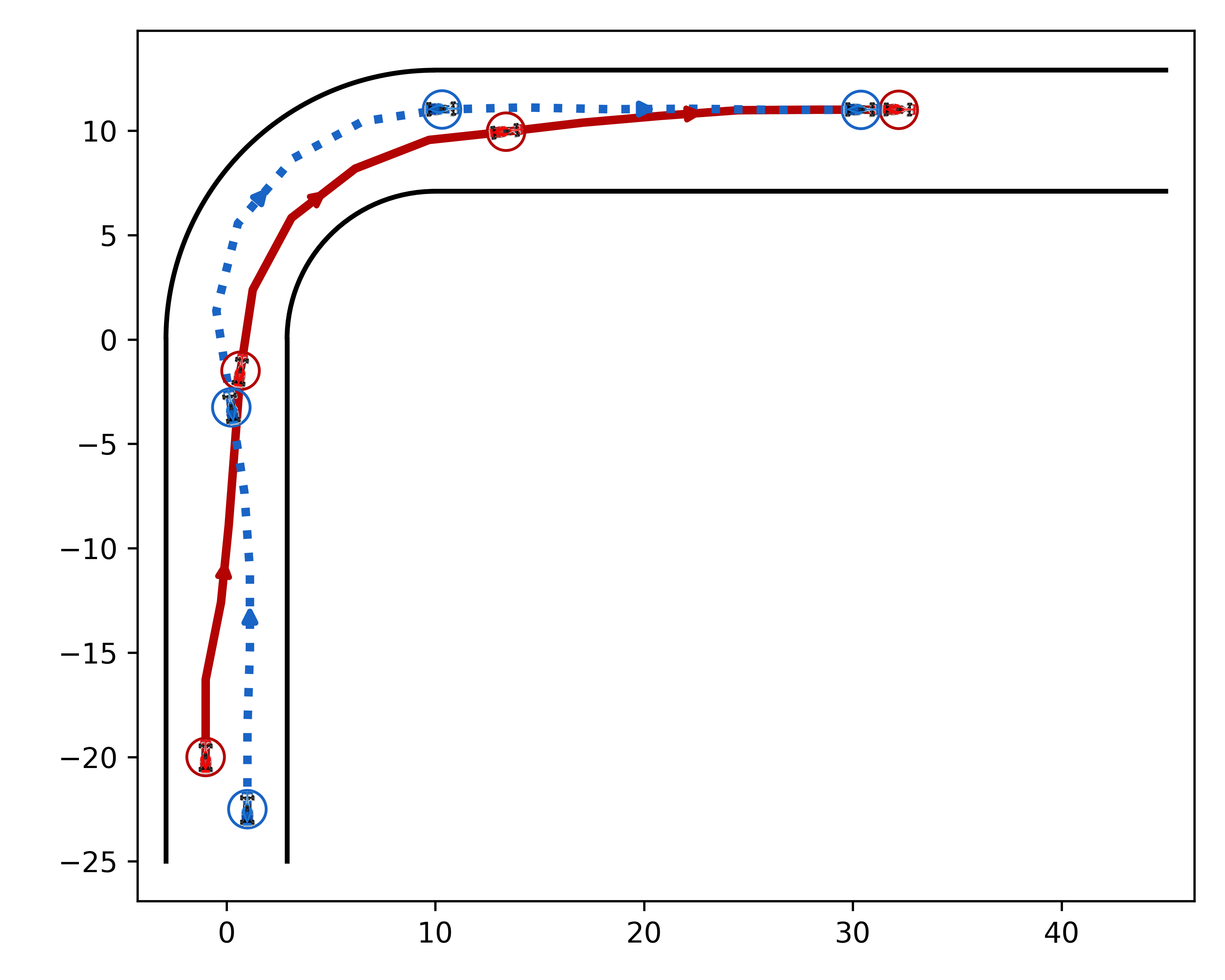}}
\subfigure[Setting 4 at corner with SPS1]{\includegraphics[scale=0.28]{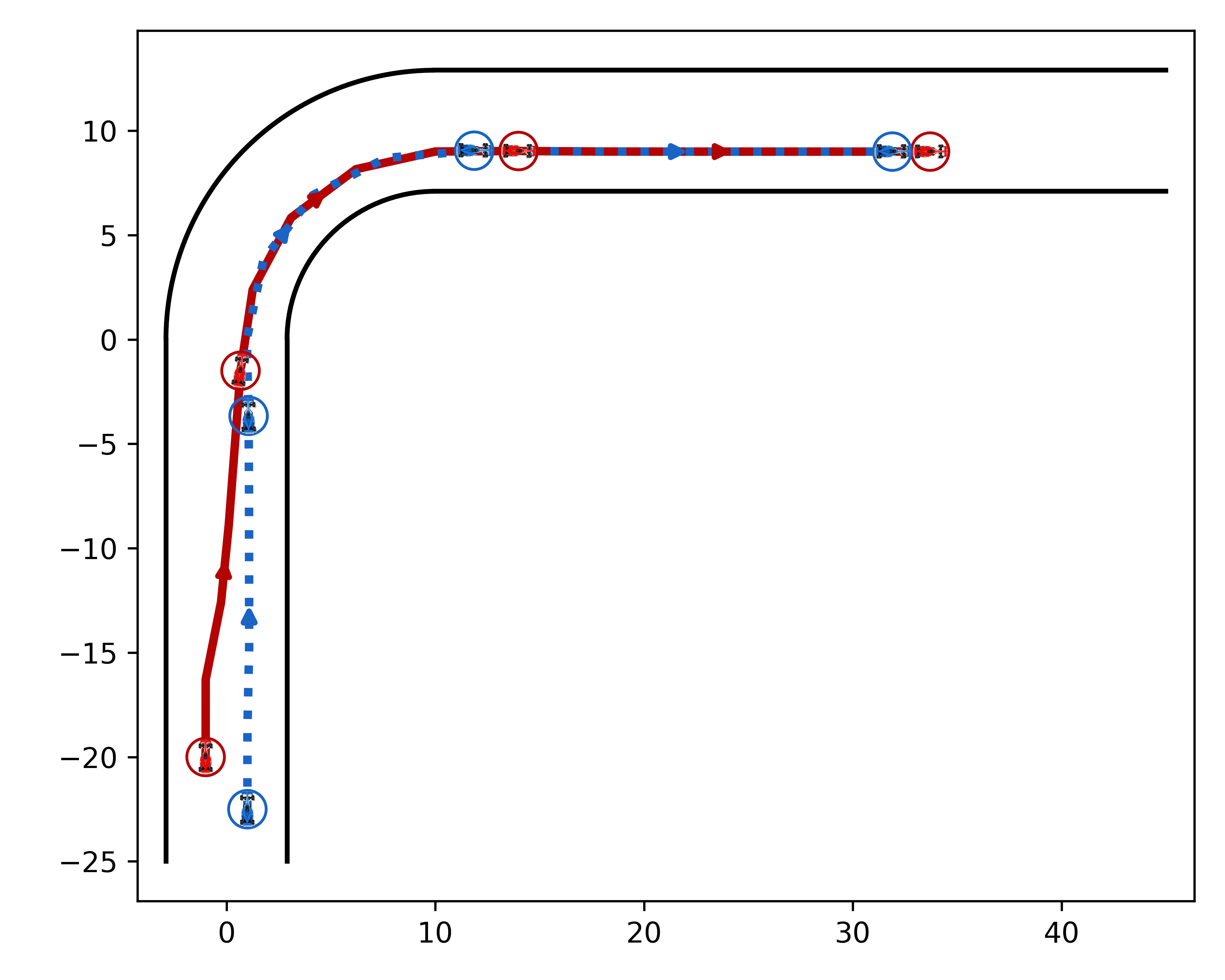}}
\caption{Simulation results at the corner under SPS1. The results are similar to the straightaway case, such that the attacker manages to perform the overtaking if and only if both players know the existence of SPS1 (as is shown in (b)). Note that results in (c) are slightly different from that in the straightaway case. After the defender switches to the inner lane for blocking, the attacker switches to the outer lane for an overtaking attempt. Since a longer distance is covered along the outer lane, the intended overtaking is delayed until both players exit the corner. Therefore, the best strategy of the defender in response is to stay in the inner lane and perform a second blocking only after both players exit the corner (compared to an immediate second blocking as in Fig. \ref{fig:strateghtSPS1}.(c)). This strategy will grant the defender longer progress along the track.}
\label{fig:cornerSPS1}
\end{figure*}

% \begin{figure*}[t]
% \centering
% \subfigure[Results with setting 1 at the corner under SPS1]{\includegraphics[width=0.4\textwidth]{figures/main_corner/4.png}}
% \hfill
% \subfigure[Results with setting 2 at the corner under SPS1]{\includegraphics[width=0.4\textwidth]{figures/main_corner/5.png}}
% \caption{Simulation results }
% \label{fig:cornerSPS1}
% \end{figure*}

\begin{figure*}[t]
\centering
\subfigure[Setting 1 on the straightaway with SPS2]{\includegraphics[scale=0.43]{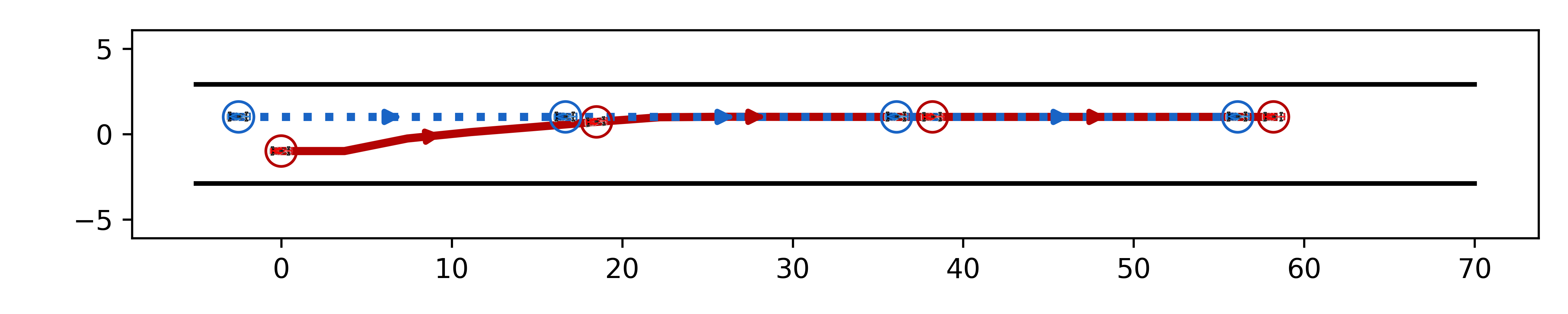}}
\subfigure[Setting 2 on the straightaway with SPS2]{\includegraphics[scale=0.43]{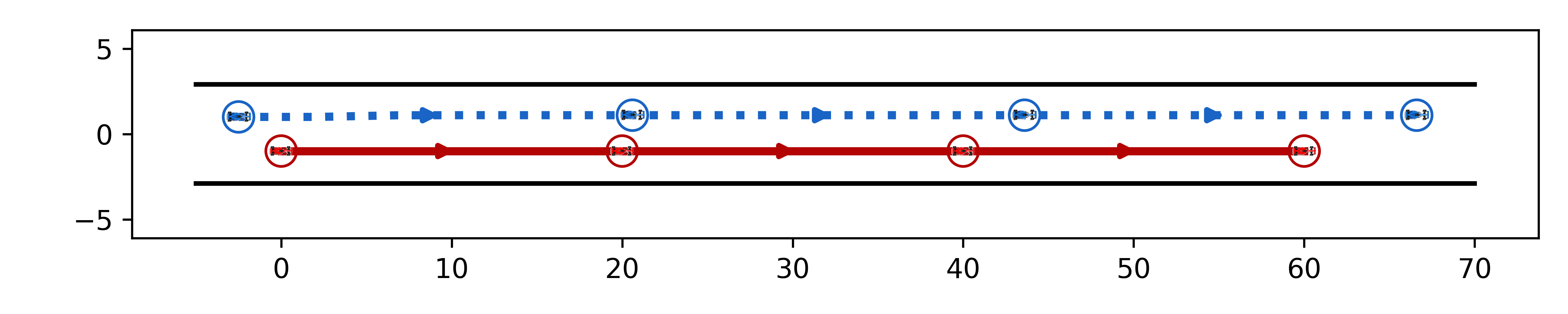}}
\subfigure[Setting 3 on the straightaway with SPS2]{\includegraphics[scale=0.43]{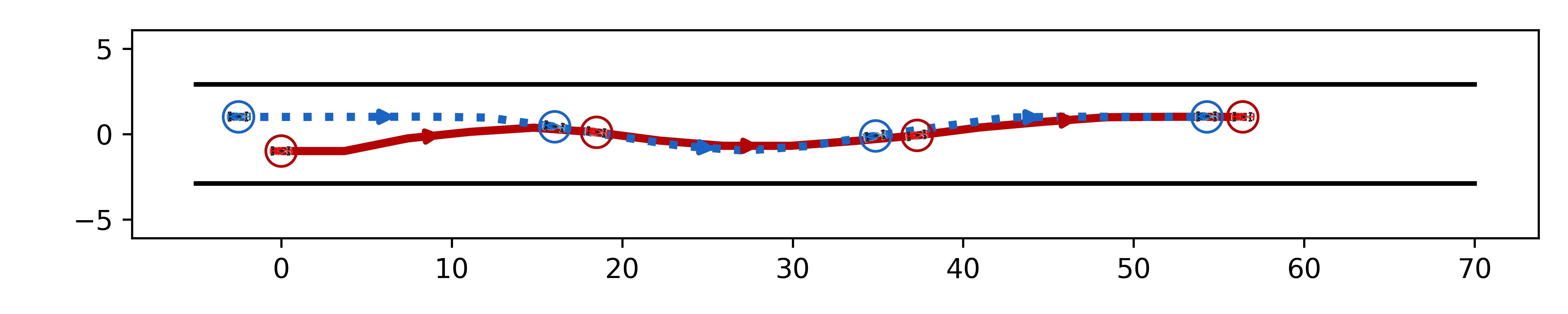}}
\subfigure[Setting 4 on the straightaway with SPS2]{\includegraphics[scale=0.43]{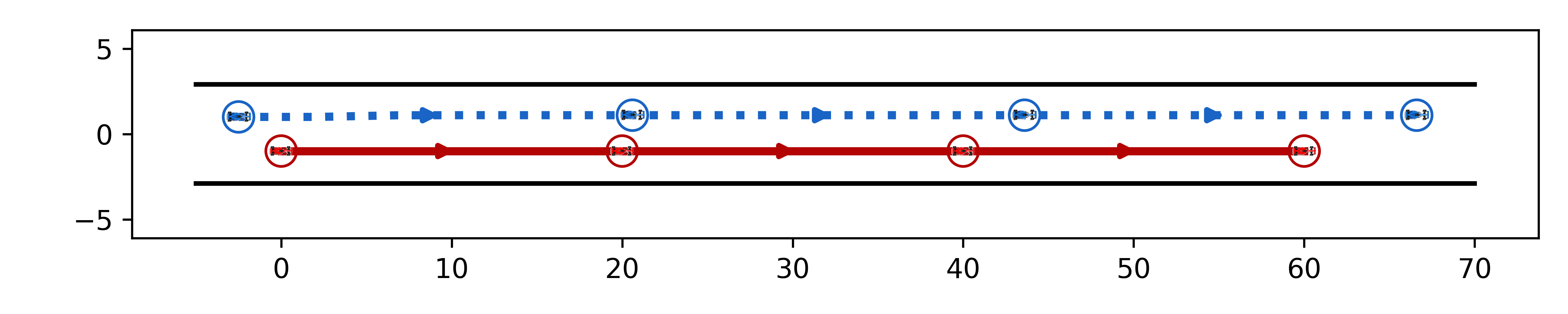}}
\caption{Simulation results on the straightaway under SPS2. In (a) where none of the players know SPS2, the defender blocks the attacker immediately. In (b) where both players know SPS2, the defender realizes that blocking is illegal under the current situation, and therefore it maintains its own lane and lets the attacker overtake. In (c) where only the attacker knows SPS2, it performs several attempts to overtake without success, since the defender continues to perform illegal blocking. In (d), although the attacker does not know the existence of SPS2, it still unintentionally utilizes SPS2 to perform a successful overtaking by keeping its own lane, which is different from the case under SPS1.}
\label{fig:SPS2}
\end{figure*}

\begin{table*}[t]
\centering
\caption{Average leading distances $\Delta \overline{x}$ and the violation rate of SPS.}
\label{tab:sps_results}
\begin{threeparttable}
\resizebox{.95\textwidth}{!}{ % Resize table to fit within text width
\renewcommand\arraystretch{1.2} % Increase row height for better readability
\begin{tabular}{cc|cccc|cccc}
    \toprule
    &  & \multicolumn{4}{c|}{Straightaway} & \multicolumn{4}{c}{Corner} \\
    &  & Setting 1 & Setting 2 & Setting 3 & Setting 4 & Setting 1 & Setting 2 & Setting 3 & Setting 4 \\ 
    \midrule
    \multirow{2}{*}{SPS1} & $\Delta \overline{x} (\textup{m})^1$ & -2.10 & 4.27 & -2.10 & -2.10 & -1.96 & 2.50 & -1.96 & -1.96 \\
    & SPS Violation Rate$^2$ & 0.04 & 0.00 & 0.94 & 0.00 & 0.20 & 0.00 & 0.88 & 0.00 \\ 
    \midrule
    \multirow{2}{*}{SPS2} & $\Delta \overline{x} (\textup{m})$ & -2.10 & 1.46 & -2.10 & -0.52 & -1.96 & 0.86 & -1.96 & -0.64 \\
    & SPS Violation Rate & 0.16 & 0.00 & 0.36 & 0.00 & 0.16 & 0.00 & 0.32 & 0.00 \\ 
    \midrule
    \multirow{2}{*}{SPS1 and SPS2} & $\Delta \overline{x} (\textup{m})$ & -2.10 & 4.78 & -2.10 & -0.52 & -1.96 & 3.20 & -1.96 & -0.54 \\
    & SPS Violation Rate & 0.18 & 0.00 & 0.94 & 0.00 & 0.28 & 0.00 & 0.88 & 0.00 \\ 
    \bottomrule
\end{tabular}
} % End resize box

\begin{tablenotes}
    \item[1] The average leading distance $(\textup{m})$ of the attacker at the end of the game. When $\Delta \overline{x} >0$, the attacker finally overtakes the defender. 
    \item[2] Percentage of frames where the defender violates SPS rules. Higher means the defender more frequently violates one of the SPS rules.
\end{tablenotes}
\end{threeparttable}
\vspace{-4mm}
\end{table*}

\subsection{Main Results}

We perform experiments on two scenarios, including a straightaway and a corner. In the straightaway scenario, the initial position of the attacker and the defender are set to be $(-2.5\,\textup{m},1\,\textup{m})$ and $(0\,\textup{m},-1\,\textup{m})$, respectively. In the corner scenario, the initial position of the attacker and the defender are set to be $(1\,\textup{m}, -22.5\,\textup{m})$ and $(-1\,\textup{m},-20\,\textup{m})$, respectively. In both scenarios, the initial velocities of the attacker and the defender are set to be $12\,\textup{m}/\textup{s}$ and $10\,\textup{m}/\textup{s}$, and the maximal velocities of both players are set to be the same as initial velocities. Qualitative results for SPS1 are shown in Fig. \ref{fig:strateghtSPS1} and Fig. \ref{fig:cornerSPS1}. From these results, we can reach the following conclusions: 1) The attacker can not perform a successful overtaking if the blocking behavior of the defender is not subject to the requirement SPS1, as is shown by Fig. \ref{fig:strateghtSPS1}(a), Fig. \ref{fig:strateghtSPS1}(c), Fig. \ref{fig:cornerSPS1}(a), and Fig. \ref{fig:cornerSPS1}(c), even if its maximal velocity is larger than the defender. This fact supports our claim that SPS is necessary for ensuring fairness in the racing competition. 2) The attacker fail to perform a successful overtaking if it does not know SPS1, as is shown by Fig. \ref{fig:strateghtSPS1}(d) and Fig. \ref{fig:cornerSPS1}(d). This fact implies the importance of knowing SPS and devising an attacking strategy based on SPS. Qualitative results for SPS2 in the corner scenario are shown in Fig. \ref{fig:SPS2}. From these results, it is clear that the first conclusion in the case of SPS1 still holds, as is shown by Fig. \ref{fig:SPS2}(a) and Fig. \ref{fig:SPS2}(c). What is different in this case is that the attacker may trigger SPS2 unintentionally to gain the leading position even if it does not know SPS2.

To further consolidate our conclusions, quantitative experiments are performed. For each scenario, we generate a set of 50 different combinations of the initial states of both players through random sampling. We then perform simulations over these intial states for the 4 settings and the 3 cases in Section V-A. For each SPS case and each setting, we measure the average leading distance of the attacker at the end of the game, $\Delta \overline{x}$, and the violation rate of the defender. Results are shown in Table II. It can be shown that under setting 2, the attacker obtains the largest leading distances under all cases, which supports our previous conclusion. Moreover, it can be seen that under SPS1, the average leading distances obtained by adopting setting 4 is the same as setting 1 and 3, which implies that SPS1 can hardly be triggered unintentionally by the attacker. The same indices for SPS2 imply that SPS2 can possibly (but not always) be triggered even if the attacker does not know the existence of SPS2. These conclusions are consistent with the ones drawn from qualitative analysis. The violation rate of SPS show that the proposed method can generate strategies that are consistent with the requirements of the SPS as long as the SPS is known by the defender (see SPS violation rate under setting 2 and 4).

\section{Conclusion} \label{Sec: conclusion}

In this paper, we study the impact of SPS on racing competitions. To incorporate SPS in racing games and generate optimal strategies for all racing vehicles with compliance of SPS rules, a novel trajectory planning scheme is introduced based on bi-level game formulation. Simulations are performed on different scenarios with different SPS rules. The results verify the effectiveness of the proposed scheme, and also highlight the importance of considering SPS in racing competition. The main limitation of this work is that vanilla MCTS is hardly scalable with respect to the number of players and the depth of the game tree. Therefore, the proposed method can only handle simple intentions of low frequency. Possible future works include combining MCTS with deep reinforcement learning to improve the performance of the proposed method, and generate the conclusion to more complicated racing scenarios and other SPS rules.

%%%%%%%%%%%%%%%%%%%%%%%%%%%%%%%%%%%%%%%%%%%%%%%%%%%%%%%%%%%%%%%%%%%%%%%%%%%%%%%%
\bibliographystyle{IEEEtran}
\bibliography{references}

\end{document}